%% file: graphPropagation_arxiv.tex
\documentclass[10pt]{IEEEtran}

\usepackage{amsthm,amsmath,amssymb}
\usepackage{cleveref}
\usepackage{mathtools,cite}
\usepackage{enumerate,subfigure,float}
\usepackage{bm}
\usepackage{stmaryrd}
\usepackage{mathbbol}
\usepackage{graphicx}
\usepackage{scrextend}
\usepackage{tikz}
\usepackage{graphicx}
\usepackage{colortbl,color}
\usepackage{psfrag}
\usepackage{algorithm}
\usepackage[noend]{algorithmic}
\usepackage{cancel}

\usetikzlibrary{positioning}

\newcommand{\Z}{\mathbb{Z}}
\newcommand{\err}{h}
\newcommand{\dleqslant}{\,\dot{\leqslant}\,}
\newcommand{\Zn}{\mathbb{Z}_n}
\newcommand{\Zl}{\mathbb{Z}_\err}
\newcommand{\Zp}{\mathbb{Z}_{\geqslant 0}}

\newcommand{\G}{\mathcal{G}}
\newcommand{\V}{\mathcal{V}}
\newcommand{\E}{\mathcal{E}}
\newcommand{\N}{\mathcal{N}}
\newcommand{\mD}{\mathcal{D}}
\newcommand{\rrarr}{\rightrightarrows}
\newcommand{\mS}{\mathcal{S}}

\newcommand{\F}{\mathcal{F}}
\newcommand{\A}{\mathcal{A}}
\newcommand{\bP}{\mathbb{P}}
\newcommand{\mP}{\mathcal{P}}
\newcommand{\grp}{\mathrm{gph}}

\newcommand{\bE}{\mathbb{E}}
\newcommand{\bI}{\mathbb{I}}

\newcommand{\mI}{\mathcal{I}}
\newcommand{\Q}{\mathcal{Q}}

\newcommand{\fR}{\mathfrak{R}}
\newcommand{\fV}{\mathfrak{V}}
\newcommand{\mJ}{\mathcal{J}}
\newcommand{\be}{\mathbb{e}}

\newcommand{\SIRM}{\mathcal{SIRM}}

\newcommand{\M}{\mathcal{M}}
\newcommand{\W}{\mathcal{W}}

\theoremstyle{theorem}
\newtheorem{thm}{Theorem}
\newtheorem{prop}{Proposition}
\newtheorem{coro}{Corollary}

\theoremstyle{definition}

\newtheorem{sass}{Assumption}

\theoremstyle{remark}
\newtheorem{ex}{Example}
\newtheorem{rem}{Remark}

\IEEEoverridecommandlockouts
\IEEEpubid{\makebox[\columnwidth]{
\begin{minipage}{\columnwidth}
\allowbreak
\copyright 2022 IEEE.  Personal use of this material is permitted.  Permission from IEEE must be obtained for all other uses, in any current or future media, including reprinting/republishing this material for advertising or promotional purposes, creating new collective works, for resale or redistribution to servers or lists, or reuse of any copyrighted component of this work in other works.
\end{minipage}
} \hspace{\columnsep}\makebox[\columnwidth]{ }}

\title{Reachability analysis in stochastic directed graphs  by reinforcement learning}

\author{Corrado Possieri, Mattia Frasca, and Alessandro Rizzo
\thanks{C. Possieri is with Istituto di Analisi dei Sistemi ed Informatica ``A. Ruberti'', Consiglio Nazionale delle Ricerche (IASI-CNR), 00185 Roma, Italy (e-mail: corrado.possieri@iasi.cnr.it).
M. Frasca is with Dipartimento di Ingegneria Elettrica Elettronica e Informatica, University of Catania, 95131 Catania, Italy,
and also with Istituto di Analisi dei Sistemi ed Informatica ``A. Ruberti'', Consiglio Nazionale delle Ricerche (IASI-CNR), 00185 Roma, Italy
(e-mail: mattia.frasca@dieei.unict.it).
A. Rizzo is with Dipartimento di Elettronica e Telecomunicazioni, Politecnico di Torino, 10129 Torino, Italy
(e-mail: alessandro.rizzo@polito.it).}
\thanks{A.R. acknowledges financial support by Compagnia di San Paolo.
M.F. acknowledges partial financial support by the Italian Ministry for Research and Education through Research
Program PRIN 2017 under Grant 2017CWMF93, project VECTORS.
C.P. acknowledges partial financial support by Regione Lazio through Research Program 
POR FESR LAZIO 2014-2020 - GRUPPI DI RICERCA 2020 under Grant GeCoWEB A0375-2020-36616,
project OPENNESS.}}

\begin{document}

\maketitle
\IEEEpubidadjcol

\begin{abstract}
We  characterize the reachability probabilities in stochastic directed graphs by means of
reinforcement learning methods. In particular, we show that the dynamics of the transition probabilities 
in a stochastic digraph can be modeled via a difference inclusion, which, in turn, can be interpreted
as a Markov decision process. Using the latter framework, we offer a methodology to design reward functions
 to provide upper and lower bounds on the reachability probabilities of a set of nodes for stochastic digraphs.
The effectiveness of the proposed technique is demonstrated by application to the diffusion of epidemic diseases 
over time-varying contact networks generated by the proximity patterns of mobile agents.
\end{abstract}


\section{Introduction}\label{sec:intro}

Earlier studies on networks widely relied on static graphs as a main analysis tool, 
that is, graphs in which the edge set does not vary in time. This modeling paradigm successfully contributed to the analysis of networked systems, even of large dimension, such as social \cite{tian2013generalized} and biological \cite{vecchio2015cortical} networks,  social psychology systems \cite{frasca2013gossips}, electrical circuits, chemical reactions \cite{possieri2018stochastic},  and industrial supply chains \cite{wagner2010assessing}.
However, static graphs fail to capture the inherent time-varying and non-deterministic nature of the relationship between networked dynamical units.
Stochastic directed graphs (digraphs) have been introduced in \cite{possieri2016weak}
to account for time-varying networked processes in which, at each time instant, the graph topology is constructed by selecting an edge set from a finite collection according to a given probability distribution. Typical systems that have been successfully modeled through this paradigm are chemical reactions \cite{possieri2018stochastic}, sensor networks \cite{possieri2016weak}, and
biological systems having non-unique behaviors \cite{possieri2017asymptotic}.

For such a class of dynamical processes, it is often of interest to quantify the reachability probabilities of a given set, which are related to the possibility of the system to assume one of the possible configurations at a certain time. This property is important to characterize the dynamical behavior of several classes of systems, such as sensor networks and power grids, as exemplified in \cite{possieri2016weak,abate2008probabilistic}.

Computational methods based on the optimization of auxiliary functions have been proposed to characterize reachability probabilities in stochastic digraphs \cite{possieri2016weak,possieri2017asymptotic,possieri2018stochastic}; however, these methods may entail a high computational burden, which can make the problem intractable when the system size increases. In this technical note, we tackle this issue by computing reachability probabilities of stochastic digraphs through an approach based on dynamic programming and reinforcement learning. Transition probabilities between configurations are modeled using difference inclusions; then, the obtained model is paralleled to a suitable Markov decision process. A reinforcement learning strategy based on reward functions constructed on the basis of the Markov decision process is then exploited to establish upper and lower bounds on the reachability properties of the stochastic digraph. Our approach is validated with an application to the diffusion of epidemic diseases on mobile populations.

\section{Preliminaries\label{sec:stocDig}}


Let $\Zn:=\{1,\dots,n \}$ and let $\wp(\Zn)$ denote the \emph{power set} of $\Zn$,
i.e., the set of all the subsets of $\Zn$.
Let $\be_i$ denote the $i$-th row of the identity matrix.
The symbol $\mathbb{1}$ denotes the vector $[\begin{array}{ccc}
1 & \cdots & 1
\end{array}]^\top$.
The symbols $\land$, $\lor$, and $\lnot$ represent the logical AND, OR, and NOT operators,
respectively. 
The symbol $\bE[\cdot]$ denotes the expected value and the symbol $\dleqslant$ denotes the entry-wise $\leqslant$ operator.
Given a matrix $M$,  $[M]_{i,j}$ indicates its $(i,j)$th entry.
Given a set $\mI\subset\Zn$, the \emph{indicator function on $\mI$} is 
$\bI_{\mI}(\zeta)=1$ if $\zeta\in\mI$, or $\bI_{\mI}(\zeta)=0$ if $\zeta\notin\mI$.

A directed graph (briefly, a \emph{digraph}) is a pair $\G:=(\V,\E)$, where $\V=\Zn$ is the \emph{vertex set}
and $\E\subset \Zn\times\Zn$ is the \emph{edge set}. 
Given $\G=(\V,\E)$ and $x\in\V$,  the \emph{in-neighborhood of $x$} and the
\emph{out-neighborhood of $x$} are, respectively, 
\begin{align*}
\N^i(x)&:=\{y\in\Z_n:(y,x)\in\E \},\\
\N^o(x)&:=\{y\in\Z_n:(x,y)\in\E \}.
\end{align*}
The \emph{out-degree} (\emph{in-degree}) of $x$ is the cardinality of $\N^o(x)$ ($\N^i(x)$).
A \emph{regular digraph} is a digraph in which all vertices have the same number of out-neighbors, i.e., all the vertices have the same out-degree. 
A regular digraph whose vertices have out-degree $k$ is called a $k$-regular digraph.
Given two digraphs $\G_1=(\V,\E_1)$ and $\G_2=(\V,\E_2)$, their \emph{union} is $\G_1\cup\G_2=(\V,\E_1\cup\E_2)$. A time-varying graph is a graph whose vertex set is constant and the edge set changes in time.

A \emph{stochastic digraph} is a time-varying digraph obtained by randomly selecting, at each  time, an element from a finite collection of digraphs. 
Formally, a stochastic digraph is defined by a triple $\bm{\G}=(\V,\bm{\E},\mu(\cdot))$,
where: 
\begin{itemize}
\item $\V:=\Zn$;
\item  $\bm{\E}:=\{\bm{\E}_s \}_{s=1}^{\err}$, $\err\in\Z_{\geqslant 0}$, with $\err < \infty$,
	such that $\G_{s}:=(\V,\bm{\E}_s)$ is a digraph for each $s\in\{1,\dots,\err\}$;
\item $\mu(\cdot)$ is a distribution function derived from an infinite sequence of independent, identically distributed (i.i.d.) random variables $\bm{w}_k:\Omega\rightarrow\{1,\dots,\err\}$, $k\in\Z_{\geqslant 0}$, defined on a given probability space $(\Omega,\F,\bP)$ \cite{fristedt2013modern}.
\end{itemize}

Note that the number $h$ of admissible topologies may be much smaller than (although possibly equal to)
$2^{\binom{n}{2}}$, which is the number of digraphs possible with $n$ vertices.

The probability $\bP(\bm{w}_k\in F):=\bP(\{\omega\in\Omega:\bm{w}_k(\omega)\in F \})$
is well defined and independent of $k\in\Zp$, for each $F\in\wp(\Z_{\err})$.
Hence, the distribution function $\mu:\wp(\Z_{\err})\rightarrow[0,1]$ is defined as
\cite[Sec. 2.1 and Sec. 11.1]{fristedt2013modern}
\begin{equation}\label{eq:probMeas}
\mu(F):=\bP(\bm{w}_k\in F ),
\end{equation}
for each $F\in\wp(\Z_{\err})$.
Note that, in this setting, there is no loss of generality in considering a
finite $\err$, since the number of edge sets $\E$ such that $(\V,\E)$ is a digraph is finite \cite{possieri2017asymptotic}.

Let us define the set-valued map $H:\Z_n\times  \Zl \rrarr \Z_n$
of the out-neighborhoods in the instantaneous digraphs as
\begin{equation}
\label{eq:mapH}
H(x,w):=\{y\in\Z_n:(x,y)\in \bm{\E}_w \}.
\end{equation}

A sequence $(\bm{\phi},\bm{w}):=\{(\bm{\phi}_k,\bm{w}_k)\}_{k=0}^K$
is a \emph{regular directed path of a stochastic digraph $\bm{\G}$ starting at $x$} if 
\begin{itemize}
\item $\bm{\phi}_0=x$;
\item $\bm{\phi}_{k+1}\in\N_{\bm{w}_k}^o(\bm{\phi}_k)$, for all $k\in\{0,\dots,K-1\}$.
\end{itemize}

A map $\bm{x}$ from $\Omega$ to sequences in $\Z_n$,
that is 
$\omega\xmapsto{\bm{x}} \{\bm{x}_k(\omega)\}_{k=0}^{\bm{K}(\omega)}$,
where $\bm{K}:\Omega\rightarrow\Z_{\geqslant 0}\cup\{\infty\}$ is
	a random variable denoting the length of the path,
 is a \emph{stochastic directed path from $x\in\Zn$} if the following properties hold:
\begin{labeling}{al}
\item [\emph{Path-wise feasibility}:] for each $\omega\in\Omega$, the sequence $\{(\bm{x}_k(\omega),
\bm{w}_k(\omega)) \}_{k=0}^{\bm{K}(\omega)}$ 
	is a regular directed path from $x$.
\item [\emph{Causal measurability}:] for each $k\in\Z_{\geqslant 0}$, the mapping $\omega\mapsto\bm{x}_{k+1}(\omega)$ is $\F_k$-measurable,
	where $(\F_0,\F_1,\F_2,\dots)$ is the minimal filtration of $\bm{w}$ \cite{fristedt2013modern}. 
\end{labeling}


According to \cite{possieri2016weak}, the set-valued mapping $H(x,w)$ given in~\eqref{eq:mapH}
satisfies the standing assumption of \cite{subbaraman2013converse,grammatico2013discrete,teel2014converse},
and hence stochastic directed paths exist and are well defined.
A stochastic directed path is \emph{maximal} if it cannot be extended and is \emph{complete} if it is maximal and $\bm{K}(\omega)=+\infty$.

Let $\mS(x)$ denote the set of all the maximal stochastic directed
paths from $x$, and, given a single stochastic directed path
$\bm{x}\in\mS(x)$, let $\grp(\bm{x})=(0,\bm{x}_0)\times (1,\bm{x}_1)\times \cdots (K,\bm{x}_K)$,
where $K\in\Zp\cup\{\infty\}$ denotes the length of the sequence $\bm{x}$.


Note that, although each stochastic directed path from $x$ is a \emph{deterministic}
function from $\Omega$ to sequences in $\Z_n$, the set $\mS(x)$ is generically not a singleton, thus making
the problem of determining reachability probabilities particularly challenging.

\section{Computation of the transition probabilities in stochastic digraphs\label{sec:transition}}

In this section, we derive a series of results that,
given a stochastic digraph $\bm{\G}$, ultimately allow to determine the 
 transition probabilities among its nodes, i.e., the probability of visiting other nodes
at the next step given the current~node. 


\subsection{Bounds for the transitions probabilities}\label{ssec:lowerupper}

Let us consider a stochastic digraph $\bm{\G}=(\V,\bm{\E},\mu(\cdot))$ and let $H:\Z_n\times  \Zl \rrarr \Z_n$ be the map of the out-neighborhoods in the instantaneous digraphs as defined in~\eqref{eq:mapH}.
The following standing assumption is made throughout this technical note.

\begin{sass}\label{sass:1}
There does not exist $(i,w)\in\Zn\times\Zl$ with $\mu(w)>0$ such that $H(i,w)=\emptyset$.
\end{sass}

\begin{rem}
\Cref{sass:1} essentially requires that no node in the stochastic digraph $\bm{\G}$ is a sink with positive probability. If this holds, then maximal stochastic directed paths are complete. On the other hand, if this does not hold, then it is possible to define an augmented stochastic digraph $\bm{\G'}$ that holds the properties of the original digraph while  satisfying \Cref{sass:1}. Namely, it suffices to add a further node $n+1$ and to let $H(n+1,w)=\{n+1\}$, for all $w\in\Zl$, and $H(i,w)=\{n+1\}$, for all $(i,w)\in\Zn\times\Zl$ such that $\mu(w)>0$ and $H(i,w)=\emptyset$. 
Stochastic paths of $\bm{\G'}$ containing the node $n+1$ represent maximal paths that are not complete. 
\end{rem}

The next theorem proves that, if a stochastic digraph satisfies \Cref{sass:1}, then it is possible to compute upper and lower bounds on the transition probabilities among its nodes. 

\begin{thm}\label{thm:bounds}
Let $\bm{\G}$ be a stochastic digraph satisfying \Cref{sass:1} and let $\bm{x}$ be a stochastic directed path of $\bm{\G}$,
i.e., $\bm{x}\in\mathcal{S}(x)$ for some $x\in\Z_n$.
For each $i,j\in\Zn$, the scalars 
\begin{subequations}
\begin{align}
\ell_{i,j} &:=\sum_{s\in\{w\in\Z_{\err}:H(i,w)= \{j\} \}}\mu(s)\in[0,1],\\
m_{i,j} &:=\sum_{s\in\{w\in\Z_{\err}:H(i,w)\cap\{j\}\neq \emptyset \}}\mu(s)\in[0,1],
\end{align}
\label{eq:matrixEnt}%
\end{subequations}
are upper and lower bounds on the transition probability $\mathbb{P}(\bm{x}_{k+1}=j\vert \bm{x}_k=i)$ from node $i$ to node $j$, 
respectively,~i.e.,
\begin{equation}\label{eq:probBounds}
\ell_{i,j} \leqslant \mathbb{P}(\bm{x}_{k+1}=j\vert \bm{x}_k=i) \leqslant m_{i,j}.
\end{equation}
\end{thm}

\begin{proof}
Let us start by considering the special case of a stochastic digraph $\bm{\G}$ for which
the digraphs $\G_s=(\V,\bm{\E}_s)$, $s=1,\dots,\err$, are $1$-regular. Then, for each $x\in\Zn$ and each $w\in\Zl$, $H(x,w)$ is a singleton, i.e., $H(x,w)=\{y\}$ for some $y\in\Zn$. It follows that the transition probabilities in $\bm{\G}$ can be represented by a Markov chain and 
\begin{multline}\label{eq:pij}
\mathbb{P}(\bm{x}_{k+1}=j\vert \bm{x}_k=i)=\sum_{s\in\{w\in\Z_{\err}:H(i,w)=\{j\} \}}\mu(s).
\end{multline}
Thus, the bounds in~\eqref{eq:probBounds} hold since $H(i,w)\cap\{j\}\neq \emptyset$ if and only if $H(i,w)=\{j\}$ under \Cref{sass:1}.

Let us now move to the more general case, where $\G_s=(\V,\bm{\E}_s)$, $s=1,\dots,\err$, may not be $1$-regular. In this case, $H(x,w)$ is a set-valued map. Let $\mI_j=\{j\}$ and consider the functions $m_j:\Zp\times \Zn\rightarrow[0,1]$ and $\ell_j:\Zp\times \Zn\rightarrow[0,1]$, defined as
$m_{j}(0,\xi):=0$ and $\ell_{j}(0,\xi):=0$,
for all $\xi\in\Z_{n}$, and, for all $(k,\xi)\in\Zp\times \Zn$,
\begin{align*}
m_{j}(k+1,\xi)&:=\sum_{w=0}^{\bar{\err}-1}\max_{g\in H(\xi,w)}\{\bI_{\mI_j}(g),m_{j}(k,g)\}\mu(w),\\
\ell_j(k+1,\xi)&:=\sum_{w=0}^{\bar{\err}-1}\min_{g\in H(\xi,w)}\{\bI_{\mI_j}(g),\bI_{\mI_j^\complement}(g)\ell_{j}(k,g)\}\mu(w),
\end{align*}
where $\mI_j^\complement:=\Z_{n}\setminus \mI_j$.
Since the set $\mI_j$ is both closed and open \cite{possieri2017asymptotic}, by \cite{subbaraman2013converse} one has that,
for all $\kappa\in\Zp$, $\kappa\geqslant 1$,
$m_{j}(\kappa,\xi) 
= \sup_{\bm{\xi}\in\mS(\xi)} \bP(\exists k\in\{1,\dots,\kappa\}:\bm{\xi}_k=j),$
and there exists $\overline{\bm{\xi}}\in\mS(\xi)$ such that the $\sup$ is attained.
In addition, by \cite{possieri2017lyapunov}, one has that 
$\ell_{j}(\kappa,\xi) 
= \inf_{\bm{\xi}\in\mS(\xi)} \bP(\exists k\in\{1,\dots,\kappa\}:\bm{\xi}_k=j)$,
and there exists $\underline{\bm{\xi}}\in\mS(\xi)$ such that the $\inf$ is attained.
This proves that~\eqref{eq:probBounds} holds with $m_{i,j}$ and $\ell_{i,j}$ given in~\eqref{eq:matrixEnt},
due to the fact that the random variables $\bm{w}_k$ are i.i.d., and that $\max_{g\in H(\xi,w)}\{\bI_{\mI_j}(g),0\}$ equals $1$ if
$H(\xi,w)\cap\{j\}\neq \emptyset$ and $0$ otherwise, whereas $\min_{g\in H(\xi,w)}\{\bI_{\mI_j}(g),0\}$ equals
$1$ if $H(\xi,w)=\{j\}$ and $0$ otherwise; see \cite[Prop.~3]{possieri2017lyapunov}.
\end{proof}

In the special case of a stochastic digraph $\bm{\G}$ with  $1$-regular
 digraphs $\G_s=(\V,\bm{\E}_s)$, $s=1,\dots,\err$, 
 we can introduce the matrix $P$ with entries $p_{i,j}=\mathbb{P}(\bm{x}_{k+1}=j\vert \bm{x}_k=i)$.
In all the other cases, rather than $P$, we consider the matrices $M$
and $L$, with entries given by the terms $m_{i,j}$ and $\ell_{i,j}$ in~\eqref{eq:matrixEnt}. 
Note that the matrix $P$ is \emph{right stochastic} \cite{gagniuc2017markov}, i.e., its rows sum to $1$, 
whereas the matrices $M$ and $L$ may not satisfy this property.

\begin{ex}\label{ex:LM}
Consider the stochastic digraph $\bm{\G}=(\V,\bm{\E},\mu(\cdot))$, where $\V:=\Z_{4}$, 
$\bm{\E}:=\{\bm{\E}_s\}_{s=1}^2$,
${\mu(\{1\})=2/3}$, ${\mu(\{2\})=1/3}$, with $\G_1
=(\V,\bm{\E}_1)$ and $\G_2=(\V,\bm{\E}_2)$ as in \Cref{fig:simple}.

\begin{figure}[htb!]
\centering
\subfigure[ $\G_1=(\V,\bm{\E}_1)$.]{
\includegraphics[width=0.25\columnwidth]{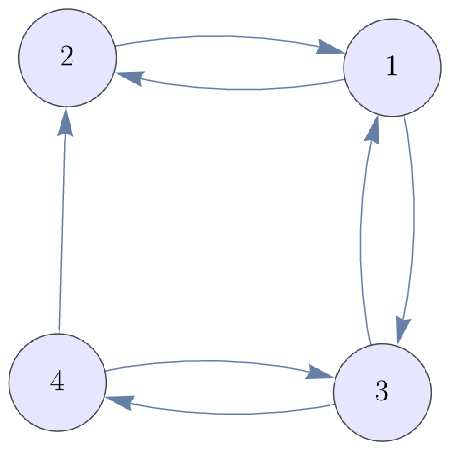}}
\qquad\qquad
\subfigure[ $\G_2=(\V,\bm{\E}_2)$.]{
\includegraphics[width=0.25\columnwidth]{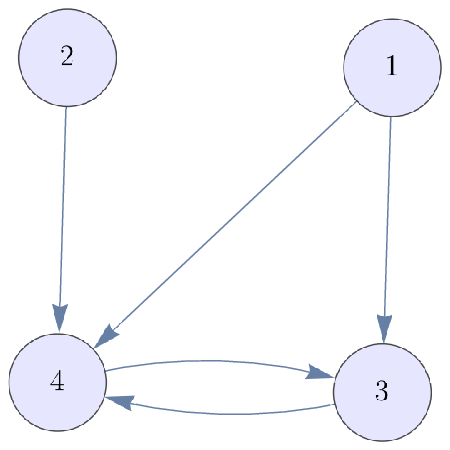}}
\caption{An example of stochastic digraph with $h=2$.\label{fig:simple}}
\end{figure}

First, we calculate the set-valued maps $H(x,w)$
of the out-neighborhoods, defined as in~\eqref{eq:mapH}, for each $x\in\Zn$ and $w\in\Zl$:
\begin{align*}
H(1,1)&=\{2,3\},& H(1,2) &= \{3,4\},&H(2,1)&=\{1\}, \\
 H(2,2) &=\{4\},&H(3,1)&=\{1,4\},& H(3,2) & = \{4\},\\
H(4,1)&=\{ 2,3\}, & H(4,2) & = \{3\}.
\end{align*}

Then, expressions~\eqref{eq:matrixEnt} are used to compute matrices $L$ and $M$ for the stochastic digraph $\bm{\G}$,
obtaining
\begin{align*}
L & =\left[
\begin{array}{cccc}
 0 & 0 & 0 & 0 \\
 \frac{2}{3} & 0 & 0 & \frac{1}{3} \\
 0 & 0 & 0 & \frac{1}{3} \\
 0 & 0 & \frac{1}{3} & 0 \\
\end{array}
\right], &
M &= \left[
\begin{array}{cccc}
 0 & \frac{2}{3} & 1 & \frac{1}{3} \\
 \frac{2}{3} & 0 & 0 & \frac{1}{3} \\
 \frac{2}{3} & 0 & 0 & 1 \\
 0 & \frac{2}{3} & 1 & 0 \\
\end{array}
\right].
\end{align*}
\end{ex}

\subsection{Dynamics of the probability vector}\label{ssec:diffIncl}

Let us now consider the probability vector 
\begin{equation*}
\rho_k=[\begin{array}{ccc}
\bP(\bm{x}_k=1) & \cdots & \bP(\bm{x}_k=n)
\end{array}],
\end{equation*}
whose $i$-th entry represents the probability of visiting the $i$-th node at  time $k$.
By \Cref{thm:bounds}, one has 
$\rho_k\,L \dleqslant \rho_{k+1} \dleqslant  \rho_k \,M$,
although the entries of $\rho_k\,L $ and $\rho_k \,M$ need not sum to $1$. 

\begin{thm}\label{thm:prob}
Let $\bm{\G}$ be a stochastic digraph satisfying \Cref{sass:1}. Then, there exist right stochastic matrices $P_1,\dots,P_\nu$ such that
the dynamics of $\rho_k$ are given by 
\begin{subequations}
\label{eq:inclProb}%
\begin{align}
P_{k} &\in \{P_1,\dots,P_\nu\}=:\mP,\label{eq:Pk+1}\\
\rho_{k+1} & = \rho_k\,P_k,
\end{align}
\end{subequations}
with domain $\mD:=\{(\rho,P)\in[0,1]^{1,n}\times \mP: \rho\,\mathbb{1}=1  \}$.
\end{thm}
\begin{proof} 
%
Consider the digraphs $\G_s=(\V,\bm{\E}_s)$, $s=1,\dots,\err$.
Each of the digraphs $\G_s$ can be decomposed into the union of a finite number of $1$-regular digraphs,  
$\G_s=:\bigcup_{i=1}^{T_s}\bar{\G}_{s,i}$, $ s=1,\dots,\err$.
Hence, letting $\bar{\G}_{s,i}=:(\V,\E_{s,i})$ and defining $\nu=\prod_{s=1}^h T_s$, consider all the  stochastic digraphs
$\bm{\G_a}=(\V,\bm{\E_a},\mu(\cdot))$, with $a=1,\ldots,\nu$, obtained letting $\bm{\E_a}=\{{(\bm{\E_a}})_s \}_{s=1}^{\err}$
and $({\bm{\E_a}})_s=\E_{s,i}$ for some $i\in\{1,\dots,T_s\}$, $s=1,\dots,\err$.
Since each $\bm{\G_a}$ satisfies the hypothesis that $H_a(x,w)=\{y\}$ for some $y\in\Zn$, for each $(x,w)\in\Zn\times\Zl$, then the transition probabilities in $\bm{\G_a}$ can be represented through a Markov chain,
whose transition matrix $P_a$ is given by~\eqref{eq:pij}. Hence, letting $P_1,\dots,P_\nu$ be these matrices, the thesis follows by the definition of stochastic directed paths.
\end{proof}

From \Cref{thm:prob}, we derive a procedure to determine $\mP$, formalized in \Cref{alg:setP}.

\begin{algorithm}[htb!]
\caption{Computation of the set $\mP$\label{alg:setP}}
\begin{algorithmic}[1]
\REQUIRE stochastic digraph $\bm{\G}$
\ENSURE set $\mP$ such that Eq.~\eqref{eq:inclProb} holds
\FOR{$s=1$ \TO $\err$}
	\STATE determine $1$-regular digraphs $\bar{\G}_{s,1},\dots,\bar{\G}_{s, T_s}$ such that 
		$\G_s:=(\V,\bm{\E}_s)=\bigcup_{i=1}^{T_s}\bar{\G}_{s,i}$ and let  $\bar{\G}_{s,i}=:(\V,\E_{s,i})$
\ENDFOR
\STATE compute $\nu=\prod_{s=1}^h T_s$
\STATE \label{step:tuples}generate the list of all possible tuples $\bm{\E_1},\dots,\bm{\E_{\nu}}$, 
$\bm{\E_a}=\{(\bm{\E_a})_s\}_{s=1}^{\err}$,
with $\bm{a}=1,\dots,\nu$, whose $s$-th element is from $\{\E_{s,1},\dots,\E_{s,T_s}\}$
\FOR{$\bm{a}=1$ \TO $\nu$}
	\STATE define the stochastic digraph $\bm{\G_a}:=(\V,\bm{\E_a},\mu(\cdot))$
	\STATE use~\eqref{eq:pij} to compute the corresponding matrix $P_a$
\ENDFOR
\RETURN $\mP:=\{P_1,\dots,P_{\nu}  \}$
\end{algorithmic}
\end{algorithm}

Note that the selection of matrices $P_1,P_2,\dots,P_\nu$ in~\eqref{eq:Pk+1}
corresponds to the choice of a random stochastic path in $\mS(x)$.
Namely, following the construction made in the proof of \Cref{thm:prob} and the definition of stochastic directed paths given in \Cref{sec:stocDig}, at each time instant one has to choose a path-wise feasible rule to select the node $\bm{x}_{k+1}$ given $\bm{x}_k$. 
The selection of the matrix $P_{k}$ in~\eqref{eq:Pk+1} corresponds exactly to the choice of such a feasible rule.

The following corollary, which provides bounds on the matrices in $\mP$, 
is a direct consequence of Theorems~\ref{thm:bounds} and~\ref{thm:prob}.

\begin{coro}\label{cor:boundsP}
Suppose that the assumptions of \Cref{thm:prob} hold and consider the matrices $M$
and $L$ with entries as in~\eqref{eq:matrixEnt}. 
Then, letting $\mP$ be defined as in \Cref{thm:prob}, one has
\begin{equation*}
L \dleqslant P \dleqslant M,\quad \forall P\in\mP.
\end{equation*}
\end{coro}


\begin{ex}\label{ex:LPM}
Consider the stochastic digraph $\bm{\G}=(\V,\bm{\E},\mu(\cdot))$ given in \Cref{ex:LM}.
Following \Cref{alg:setP}, the digraphs $\G_1$ and $\G_2$ 
given in \Cref{fig:simple} can be decomposed
into the union of eight and two $1$-regular digraphs, respectively. Thus, we have $T_1=8$, $T_2=2$, and $\nu=16$.
 Considering the stochastic digraphs $\bm{\G_a}=(\V,\bm{\E_a},\mu(\cdot))$, $a=1,\ldots,16$, obtained letting
$\bm{\E_a}=\{\E_{1,i},\E_{2,j} \}$, $i=1,\dots,8$, $j=1,2$, using~\eqref{eq:pij},
one has 
\begin{multline*}
\mP=\left\{
\left[\begin{smallmatrix}
 0 & \frac{2}{3} & \frac{1}{3} & 0 \\
 \frac{2}{3} & 0 & 0 & \frac{1}{3} \\
 \frac{2}{3} & 0 & 0 & \frac{1}{3} \\
 0 & \frac{2}{3} & \frac{1}{3} & 0 \\
\end{smallmatrix}\right],
\left[\begin{smallmatrix}
 0 & \frac{2}{3} & 0 & \frac{1}{3} \\
 \frac{2}{3} & 0 & 0 & \frac{1}{3} \\
 \frac{2}{3} & 0 & 0 & \frac{1}{3} \\
 0 & \frac{2}{3} & \frac{1}{3} & 0 \\
\end{smallmatrix}\right],
\left[\begin{smallmatrix}
 0 & \frac{2}{3} & \frac{1}{3} & 0 \\
 \frac{2}{3} & 0 & 0 & \frac{1}{3} \\
 \frac{2}{3} & 0 & 0 & \frac{1}{3} \\
 0 & 0 & 1 & 0 \\
\end{smallmatrix}\right],
\left[\begin{smallmatrix}
 0 & \frac{2}{3} & 0 & \frac{1}{3} \\
 \frac{2}{3} & 0 & 0 & \frac{1}{3} \\
 \frac{2}{3} & 0 & 0 & \frac{1}{3} \\
 0 & 0 & 1 & 0 \\
\end{smallmatrix}\right],\right.\\
\left[\begin{smallmatrix}
 0 & \frac{2}{3} & \frac{1}{3} & 0 \\
 \frac{2}{3} & 0 & 0 & \frac{1}{3} \\
 0 & 0 & 0 & 1 \\
 0 & \frac{2}{3} & \frac{1}{3} & 0 \\
\end{smallmatrix}\right],
\left[\begin{smallmatrix}
0 & \frac{2}{3} & 0 & \frac{1}{3} \\
 \frac{2}{3} & 0 & 0 & \frac{1}{3} \\
 0 & 0 & 0 & 1 \\
 0 & \frac{2}{3} & \frac{1}{3} & 0 \\
\end{smallmatrix}\right],
\left[\begin{smallmatrix}
 0 & \frac{2}{3} & \frac{1}{3} & 0 \\
 \frac{2}{3} & 0 & 0 & \frac{1}{3} \\
 0 & 0 & 0 & 1 \\
 0 & 0 & 1 & 0 \\
\end{smallmatrix}\right],
\left[\begin{smallmatrix}
 0 & \frac{2}{3} & 0 & \frac{1}{3} \\
 \frac{2}{3} & 0 & 0 & \frac{1}{3} \\
 0 & 0 & 0 & 1 \\
 0 & 0 & 1 & 0 \\
\end{smallmatrix}\right],\\
\left[\begin{smallmatrix}
 0 & 0 & 1 & 0 \\
 \frac{2}{3} & 0 & 0 & \frac{1}{3} \\
 \frac{2}{3} & 0 & 0 & \frac{1}{3} \\
 0 & \frac{2}{3} & \frac{1}{3} & 0 \\
\end{smallmatrix}\right],
\left[\begin{smallmatrix}
 0 & 0 & \frac{2}{3} & \frac{1}{3} \\
 \frac{2}{3} & 0 & 0 & \frac{1}{3} \\
 \frac{2}{3} & 0 & 0 & \frac{1}{3} \\
 0 & \frac{2}{3} & \frac{1}{3} & 0 \\
\end{smallmatrix}\right],
\left[\begin{smallmatrix}
 0 & 0 & 1 & 0 \\
 \frac{2}{3} & 0 & 0 & \frac{1}{3} \\
 \frac{2}{3} & 0 & 0 & \frac{1}{3} \\
 0 & 0 & 1 & 0 \\
\end{smallmatrix}\right],
\left[\begin{smallmatrix}
 0 & 0 & \frac{2}{3} & \frac{1}{3} \\
 \frac{2}{3} & 0 & 0 & \frac{1}{3} \\
 \frac{2}{3} & 0 & 0 & \frac{1}{3} \\
 0 & 0 & 1 & 0 \\
\end{smallmatrix}\right],\\
\left.\left[\begin{smallmatrix}
 0 & 0 & 1 & 0 \\
 \frac{2}{3} & 0 & 0 & \frac{1}{3} \\
 0 & 0 & 0 & 1 \\
 0 & \frac{2}{3} & \frac{1}{3} & 0 \\
\end{smallmatrix}\right],
\left[\begin{smallmatrix}
 0 & 0 & \frac{2}{3} & \frac{1}{3} \\
 \frac{2}{3} & 0 & 0 & \frac{1}{3} \\
 0 & 0 & 0 & 1 \\
 0 & \frac{2}{3} & \frac{1}{3} & 0 \\
\end{smallmatrix}\right],
\left[\begin{smallmatrix}
 0 & 0 & 1 & 0 \\
 \frac{2}{3} & 0 & 0 & \frac{1}{3} \\
 0 & 0 & 0 & 1 \\
 0 & 0 & 1 & 0 \\
\end{smallmatrix}\right],
\left[\begin{smallmatrix}
 0 & 0 & \frac{2}{3} & \frac{1}{3} \\
 \frac{2}{3} & 0 & 0 & \frac{1}{3} \\
 0 & 0 & 0 & 1 \\
 0 & 0 & 1 & 0 \\
\end{smallmatrix}\right]
\right\}.
\end{multline*}
Note that all the matrices in $\mP$ are right stochastic and that their entry-wise minimum and maximum
equal the corresponding entry of $L$ and $M$, respectively. 
\end{ex}

\subsection{Stochastic digraphs as Markov decision processes\label{ssec:MDP}}

So far, we have shown that the properties of stochastic digraphs can be analyzed by studying the corresponding difference inclusion~\eqref{eq:inclProb}.
Note that such an inclusion can be interpreted as a Markov decision process \cite{sutton2018reinforcement} with
action space $\A=\{1,\dots,\nu\}$ and with
\begin{equation}\label{eq:upProb}
p(j|i,a):=\bP(\bm{x}_{k+1}=j|\bm{x}_k=i,\bm{a}_k=a)=[P_a]_{i,j}.
\end{equation}
In this context, the action $a\in\A$ 
represents the selection of the rule employed to select the node $\bm{x}_{k+1}$ given $\bm{x}_k$.
However, computing the transition matrices of a large stochastic digraph via \Cref{alg:setP} can be computationally demanding as, at 
step~\ref{step:tuples} of this algorithm, one has to generate the set of all the tuples of the possible edge sets.

\begin{rem}\label{rem:simple}
The transition probabilities $\bP(\bm{x}_{k+1}=j|\bm{x}_k=i,\bm{a}_k=a)$
can be computed more efficiently letting the action space  depend on the current value of the state.
Given a node $i$, consider the sets $H(i,w)$ with $w=1,\ldots,s$, representing all the possible sets of out-neighbors that can be reached from $i$. Then, 
the set of all the path-wise feasible rules to update node $i$, labeled using the index $a_i$, consists of a node from $H(i,1)$, one from $H(i,2)$ and so on, up to $H(i,s)$. 
The number of 
path-wise feasible rules that can be used to update node $i$ 
is then determined by all the possible tuples that can be generated in this way and is, therefore, given by $\nu_i=\prod_{w=1}^s|H(i,w)|$. 
This corresponds to the cardinality of the local action space from $i$, i.e., $\A(i):=\{1,\dots,\nu_i\}$. 
Now, let us indicate the elements of $H(i,w)$ as $\{j_{w,1},\dots,j_{w,R_w}\}$ where $R_w=|H(i,w)|$, and for each $a_i\in \A(i)$ consider the set $(\bm{\mJ_i})_{a_i}:=\{\xi_{1,a_i},\dots,\xi_{\err,a_i}\}$, representing the $a_i$-th tuple associated to the 
path-wise feasible update rule $a_i$, then  
\begin{equation*}
\bm{\mJ_i}:=\{\{j_{1,1},\dots,j_{\err,1}\},\dots,\{j_{1,R_1},\dots,j_{\err,R_\err}\} \}
\end{equation*}
represents the set of all tuples whose $w$-th element is selected from $H(i,w)$. 
Thus, the transition probabilities $p(j|i,a)=\bP(\bm{x}_{k+1}=j|\bm{x}_k=i,\bm{a}_k=a)$ for $a\in\A(i)$
are given by
\begin{equation}\label{eq:localTrans}
p(j|i,a)=\sum_{w\in\{s\in\Zl:\xi_{s,a}=j \}}\mu(w).
\end{equation}

Note that there are two clear advantages of defining $\A(i)$ and $p(j|i,a)$ as in~\eqref{eq:localTrans} rather
than as in~\eqref{eq:upProb}:
firstly, the set $\bm{\mJ_i}$ can be generated by just considering the value attained by the set-valued map $H(x,w)$ for 
$x=i$ and $w\in\Zl$ and hence the probabilities $p(j|i,a)$ can be determined online without explicitly computing 
the matrices $P_1,\dots,P_\nu$. Secondly, the action space $\A(i)$ is smaller than that of the set $\A$.
Namely, its cardinality is $\nu_i=\prod_{w=1}^s|H(i,w)|$, which is in general much smaller than $\nu$.
\end{rem}

The following algorithm summarizes the steps of the procedure given in \Cref{rem:simple}.

\begin{algorithm}[htb!]
\caption{Computation of the function $p(j|i,a)$\label{alg:probs}}
\begin{algorithmic}[1]
\REQUIRE the map of the out-neighborhoods $H(i,w)$ and the probability distribution function $\mu(\cdot)$
\ENSURE the transition probability $p(j|i,a)=\bP(\bm{x}_{k+1}=j|\bm{x}_k=i,\bm{a}_k=a)$ and the action space $\A(i)$
\STATE generate the list $\bm{\mJ_i}$
of all tuples whose $w$-th element is from $\{j_{w,1},\dots,j_{w,R_w}\}:=H(i,w)$ 
\STATE calculate $\nu_i$ as $\nu_i=\prod_{w=1}^s|H(i,w)|$ 
\STATE letting $(\bm{\mJ_i})_a:=(\xi_{1,a},\dots,\xi_{\err,a})$ be the $a$-th tuple in $\bm{\mJ_i}$, compute
	$p(j|i,a)$ as in \eqref{eq:localTrans}
\RETURN $p(j|i,a)$ and $\A(i):=\{1,\dots,\nu_i\}$ 
\end{algorithmic}
\end{algorithm}


\begin{ex}\label{ex:probs}
Consider the stochastic digraph $\bm{\G}$ given in Examples~\ref{ex:LM} and~\ref{ex:LPM}. The set-valued map $H$ is that reported in~\Cref{ex:LPM}. Using \Cref{alg:probs}, we determine the  action spaces
$\A(1)=\{1,2,3,4\}$,
$\A(2)  = \{1\}$,
$\A(3)  = \{1,2\}$,
$\A(4)  = \{1,2\}$,
and the sets $\bm{\mJ_1},\dots,\bm{\mJ_4}$, given by
\begin{gather*}
\bm{\mJ_1}=\{\{2,3\},\{2,4\},\{3,3\},\{3,4\}\},\quad
\bm{\mJ_2}=\{\{1,4\}\},\\
\bm{\mJ_3}=\{\{1,4\},\{4,4\}\},\quad
\bm{\mJ_4}=\{\{2,3\},\{3,4\}\}.
\end{gather*}
 Note that the action spaces $\A(i)$, $i=1,\ldots,4$, have maximum cardinality $4$, whereas  $\mP$ has cardinality $16$. 
Then, by using~\eqref{eq:localTrans},
one obtains the transition probabilities given in \Cref{tab:trans}.

\begin{table}[htb!]
\centering
\caption{Transition probabilities in $\bm{\G}$.\label{tab:trans}}
{\renewcommand{\arraystretch}{1.2}
\renewcommand{\tabcolsep}{4pt}
\begin{tabular}{ccc|c}
 $i$ & $a$ & $j$ & $p(j|i,a)$ \\
 \hline
 1 & 1 & 2 & $\frac{2}{3}$ \\
 1 & 1 & 3 & $\frac{1}{3}$ \\
 1 & 2 & 2 & $\frac{2}{3}$ \\
 1 & 2 & 4 & $\frac{1}{3}$ \\
 1 & 3 & 3 & 1 
\end{tabular}
 }
 \hfill
{\renewcommand{\arraystretch}{1.2}
\renewcommand{\tabcolsep}{4pt}
\begin{tabular}{ccc|c}
 $i$ & $a$ & $j$ & $p(j|i,a)$ \\
 \hline
1 & 4 & 3 & $\frac{2}{3}$ \\
 1 & 4 & 4 & $\frac{1}{3}$ \\
  2 & 1 & 1 & $\frac{2}{3}$ \\
  2 & 1 & 4 & $\frac{1}{3}$ \\
  3 & 1 & 1 & $\frac{2}{3}$ 
\end{tabular}
 }
\hfill
{\renewcommand{\arraystretch}{1.2}
\renewcommand{\tabcolsep}{4pt}
\begin{tabular}{ccc|c}
 $i$ & $a$ & $j$ & $p(j|i,a)$ \\
 \hline
  3 & 1 & 4 & $\frac{1}{3}$ \\
  3 & 2 & 4 & 1 \\
  4 & 1 & 2 & $\frac{2}{3}$ \\
  4 & 1 & 3 & $\frac{1}{3}$ \\
  4 & 2 & 3 & 1 
\end{tabular}
 }
\end{table}
\end{ex}

\begin{rem}\label{rem:causal}
The hypothesis of causal measurability is fundamental to establish the equivalence 
between stochastic directed paths and Markov decision processes.
In fact, if such an assumption about stochastic paths is removed, i.e., 
$\bm{x}_{k+1}(\omega)$ is allowed to depend on $\bm{w}_{k+1}(\omega)$, then degenerate behavior may occur as shown
in \cite[Ex.~5]{possieri2017asymptotic}. In particular, it can be  argued that removing this requirement
corresponds to a non-causal selection of the action in the Markov decision process.
\end{rem}

\section{Analysis of the weak and strong reachability in stochastic digraphs\label{sec:MDP}}

When dealing with stochastic systems with non-unique solutions, such as stochastic digraphs, one of the main
challenges is to compute upper and lower bounds on the probability that a stochastic solution visits a given set.
In the literature, such a task is usually carried out by means of auxiliary Lyapunov functions whose expected
value is decreasing along the solutions of the system (see, e.g., \cite{subbaraman2013converse,possieri2017lyapunov,teel2013matrosov}
for systems with discrete state space and \cite{possieri2017asymptotic,possieri2016weak} for those with
continuous state space). Leveraging the correspondence between stochastic digraphs and Markov decision processes
established in \Cref{ssec:MDP}, the main goal of this section is to use reinforcement learning to compute upper and lower bounds on the probability 
that a path in the stochastic digraph visits a given set of nodes \cite{sutton2018reinforcement}.
Namely,  considering the
Markov decision process~\eqref{eq:localTrans}, where the state correspond to a node of the stochastic digraph and the action
correspond to a path-wise feasible rule to extend stochastic paths, the main objective of this section is to define the rewards
\begin{equation}\label{eq:funr}
r(i,a,j):=\bE(\bm{r}_{k+1}|\bm{x}_k=i,\bm{x}_{k+1}=j,\bm{a}_k=a)
\end{equation}
to determine path-wise feasible rules that either maximize or minimize the probability of visiting a given set of nodes.
In this way, the reachability properties of the stochastic digraph $\bm{\G}$ may be characterized. 
In fact, since the action $a\in\A(i)$ corresponds to the selection of an update rule for the stochastic path, selecting 
the actions that maximize a suitably defined reward $r(i,a,j)$ yields a path in $\mS(x)$ that maximizes a given probability function. 
In particular, in \Cref{ssec:reach}, we show how to define the function $r(i,a,j)$  to obtain the weak reachability properties of
the stochastic digraph $\bm{\G}$, and in \Cref{ssec:recurr} we discuss how to define $r(i,a,j)$
 to determine the strong recurrence probabilities of the stochastic digraph $\bm{\G}$.

First, we need to introduce some additional concepts borrowed from the dynamic programming framework \cite{sutton2018reinforcement}.
Given the Markov decision process~\eqref{eq:localTrans} and the rewards~\eqref{eq:funr}, let $\pi(a|x)$ be a policy that maps each state $x\in\Zn$ to the probability of taking action $a\in\A(x)$ when the system is in the state $x$. We define the \emph{state-value function for $\pi$}
\begin{equation*}
v_{\pi}(k,x):=\bE_\pi\left[ \sum_{\tau=0}^{k-1}\bm{r}_{\tau+1} \biggr| \bm{x}_0=x\right],
\end{equation*}
which is the expected reward gathered by using $\pi$,  given that the sequence $\bm{x}=\{x_\tau\}_{\tau=0}^k$
is initialized at $\bm{x}_0=x$. Letting
\begin{equation*}
p(j,r|i,a)=\bP(\bm{x}_{k+1}=j,\bm{r}_{k+1}=r|\bm{x}_k=i,\bm{a}_k=a)
\end{equation*}
be the probability that the next state and reward are $j$ and $r$, respectively, given that the current state and action
are $i$ and $a$, respectively, by the Bellman's principle of optimality \cite{sutton2018reinforcement} we have that
the function $v_{\pi}(k,x)$ satisfies
\begin{equation}\label{eq:Bellman}
v_{\pi}(k,x) 
=\sum_{a,j,r} \pi(a|x)p(j,r|i,a)\left(r+v_{\pi}(k-1,j)\right).
\end{equation}
Hence, observing that 
$v_{\pi}(1,x)=\sum_{a,j,r} \pi(a|x)p(j,r|i,a)\,r$,
the Bellman equation~\eqref{eq:Bellman} can be used to compute the values attained by the 
state-value function for $\pi$.

Let $k$ be fixed. A policy $\pi$ is defined to be better than or equal to another policy $\pi'$ if
$v_\pi(k,x)\geq v_{\pi'}(k,x)$ for all $x\in\Zn$. By \cite{sutton2018reinforcement}, 
an optimal policy $\pi_\star$, i.e., a policy that is better than all the others,  always exists. 
Although the optimal policy does not need to be unique,
all the optimal policies share the same state-value function, called the 
\emph{optimal state-value function}, which is defined as
$v_{\star}(k,x):=\max_{\pi}v_{\pi}(k,x)$.


\subsection{Weak reachability\label{ssec:reach}}
Let a set $\Q\subset \Z_n$, an integer $k\in\Z_{\geqslant 0}$, and a vertex $x\in\Z_n$ be given. Consider a random
directed path $\bm{x}\in\mS(x)$. The condition $\grp(\bm{x})\cap (\{1,\dots,k\}\times \Q)\neq \emptyset$ requires that 
the random path $\bm{x}$ reaches the set $\Q$ in a number of steps less than or equal to $k$. Hence, 
 given $\Q\subset\Z_n$, $k\in\Z_{>0}$, and $\xi\in\Z_n$, the
\emph{weak reachability probability} are defined as
\begin{equation*}
\fR_{\Q}(k,x) :=\sup_{\bm{x}\in\mS(x)}\bP(\grp(\bm{x})\cap (\{1,\dots,k\}
\times \Q) \neq \emptyset).
\end{equation*}
The value $\fR_{\Q}(k,x)$ represents the probability that a stochastic directed path starting at $x$ that visits the set $\Q$ in $k$ or less steps exists.
Given the stochastic digraph $\bm{\G}=(\V,\bm{\E},\mu(\cdot))$ and a set $\Q\subset \Zn$, let us now introduce an augmented stochastic digraph $\check{\bm{\G}}$, obtained by: i) adding two additional states, $n+1$ and $n+2$; ii) substituting in $\bm{\E}_s$, $s=1,\dots,r$,
 the edges  $(i,j)$ such that $j\in\Q$ with new edges $(i,n+1)$;
and iii) adding the edges $(n+1,n+2)$ and $(n+2,n+2)$ to all $\bm{\E}_s$, $s=1,\dots,r$.
Then, the next proposition holds.

\begin{prop}\label{prop:Wreach}
Consider a stochastic digraph $\bm{\G}$ satisfying \Cref{sass:1} and its corresponding augmented stochastic digraph $\check{\bm{\G}}$.
Let $\check{p}(j|i,a)$ the transition probabilities for $\check{\bm{\G}}$, calculated by~\eqref{eq:localTrans} and let us fix the reward function
$r(i,a,j)=\bI_{\{n+1\}}(j)$ so that
$\bm{r}_{k+1}=\bI_{\{n+1\}}(\bm{x}_{k+1})$. 
Then,  letting $v_\star$ be the optimal state-value function corresponding to such a selection of $r$,
for all $k\in\Zp$ and all $x\in\Zn$, one has
\begin{equation}\label{eq:RQ}
\fR_{\Q}(k,x)=v_{\star}(k,x).
\end{equation}
\end{prop}



\begin{proof}
Let $\bm{\zeta}$ be a random directed path of the stochastic digraph $\bm{\G}$.
By \cite{teel2013matrosov},  we have that
$\fR_{\Q}(k,x) =\sup_{\bm{\zeta}\in\mS(x)}\bE\left[\max_{j\in\{1,\dots,k\}} \bI_{\Q}(\bm{\zeta}_j)\right]$.
Since $\check{p}(n+2|n+1,a)=1$ for all $a\in\A(n+1)$ and $\check{p}(n+2|n+2,a)=1$ for all $a\in\A(n+2)$ by
the definition of $\check{\bm{\G}}$, and taking into account that $\bm{r}_{\tau+1}=\bI_{\{n+1\}}(\bm{x}_{\tau+1})$,
we obtain
\begin{align*}
\sum_{\tau=0}^{k-1}\bm{r}_{\tau+1}&=\sum_{\tau=0}^{k-1}\bI_{\{n+1\}}(\bm{x}_{\tau+1})\\
&=\begin{cases}
1,&\text{if }\exists \tau\in\{1,\dots,k\}\text{ s.t. }\bm{x}_{\tau}=n+1,\\
0, & \text{otherwise}.
\end{cases}
\end{align*}
Therefore, since,
by the definition of $\check{\bm{\G}}$, 
one has that $\bm{x}_{\tau}=n+1$ if and only if the corresponding $\bm{\zeta}_{\tau}\in\Q$,
it results that
$\bE[\sum_{\tau=0}^{k-1}\bm{r}_{k+1}]=\bE[\max_{j\in\{1,\dots,k\}} \bI_{\Q}(\bm{\zeta}_j)]$.
Hence, the statement follows by \Cref{thm:prob} and by the definition of the optimal state-value function $v_\star$.
\end{proof}

The bound given in~\eqref{eq:RQ} is tight. In fact, by \cite[Prop.~4]{subbaraman2013converse},
for each $(k,x)\in\Zp\times \Z_n$, there exists $\overline{\bm{x}}\in\mS(x)$ such that
$\fR_{\Q}(k,x)=\bP(\exists \kappa\in\{1,\dots,k\}:\overline{\bm{x}}_\kappa\in\Q)$, i.e.,
there exists a stochastic path that attains the probability given in~\eqref{eq:RQ}.


\subsection{Strong recurrence\label{ssec:recurr}}

Under \Cref{sass:1}, given $\Q\subset\Z_n$, $k\in\Z_{>0}$, and $\xi\in\Z_n$,  the
\emph{strong recurrence probabilities} are defined as
\begin{equation*}
\fV_{\Q}(k,x) :=\inf_{\bm{x}\in\mS(x)}\bP(\grp(\bm{x})\cap (\{1,\dots,k\}
\times \Q) \neq \emptyset).
\end{equation*}
The value $\fV_{\Q}(k,x)$ represents the probability that all the stochastic directed paths starting at $x$ visit the set $\Q$ in $k$ or less steps.
As for the weak reachability probabilities, this function can be determined by computing the
optimal value function of a Markov decision process. 


\begin{prop}\label{prop:Wrec}
Consider a stochastic digraph $\bm{\G}$ satisfying \Cref{sass:1} and its corresponding augmented stochastic digraph $\check{\bm{\G}}$.
Let $\check{p}(j|i,a)$ the transition probabilities for $\check{\bm{\G}}$, calculated by~\eqref{eq:localTrans} and let us fix the reward function
$r(i,a,j)=-\bI_{\{n+1\}}(j)$, so that
$\bm{r}_{k+1}=-\bI_{\{n+1\}}(\bm{x}_{k+1})$. 
Then,  letting $v_\star$ be the optimal state-value function corresponding to such a selection of $r$,
for all $k\in\Zp$ and all $x\in\Zn$, one has
\begin{equation}\label{eq:VQ}
\fV_{\Q}(k,x)=-v_{\star}(k,x).
\end{equation}
\end{prop}

\begin{proof}
Let $\bm{\zeta}$ be a random directed path of the stochastic digraph $\bm{\G}$.
By \cite{subbaraman2013converse}, we have that
$\fV_{\Q}(k,x)=1-\sup_{\bm{\zeta}\in\mS(x)}\bE\left[\prod_{j=1}^{k} \bI_{\Z_n\setminus \Q}(\bm{\zeta}_j) \right]$.
Hence, by considering that
\begin{equation*}
\prod_{j=1}^{k} \bI_{\Z_n\setminus \Q}(\bm{\zeta}_j)=\begin{cases}
0,&\text{if }\exists \tau\in\{1,\dots,k\}\text{ s.t. }\bm{x}_{\tau}\in\Q,\\
1, & \text{otherwise},
\end{cases}
\end{equation*}
we have that $\prod_{j=1}^{k} \bI_{\Z_n\setminus \Q}(\bm{\zeta}_j) = 1 - \max_{j\in\{1,\dots,k\}} \bI_{\Q}(\bm{\zeta}_j)$.
Therefore, the statement follows by arguments similar to those used to prove \Cref{prop:Wreach}.
\end{proof}

As for~\eqref{eq:RQ}, the bound given in~\eqref{eq:VQ} is tight: by \cite[Prop.~2]{possieri2017lyapunov},
for each $(k,x)\in\Zp\times \Z_n$, there exists $\underline{\bm{x}}\in\mS(x)$ such that
$\fV_{\Q}(k,x)=\bP(\exists \kappa\in\{1,\dots,k\}:\underline{\bm{x}}_\kappa\in\Q)$, i.e.,
there exists a stochastic path that attains the probability~\eqref{eq:VQ}.

The results given in Propositions~\ref{prop:Wreach} and~\ref{prop:Wrec} are
not surprising in light of the definition of optimal value functions, observing that
 node $n+2$ is terminal, and that the only nonzero reward is gathered transitioning to such a node.
Furthermore, these results are strongly linked with \cite[Thm.~1 and Thm.~2]{abate2008probabilistic}.
In particular, the recursions in these statements to design optimal value functions
are similar to those given in~\eqref{eq:matrixEnt}.
However, while the objective in~\cite{abate2008probabilistic} is to design the control input for a class of hybrid systems 
to ensure safety, here we aim at determining the maximal and minimal probabilities of visiting a given set
for stochastic digraphs. The main objective of this technical note, in fact, is to show that the reachability properties of these discrete-time
stochastic systems with non-unique solutions can be characterized via dynamic programming.
Further, in the case that the dimension of the problem to be solved is too large,
we can resort to stochastic optimization methods, such as the \emph{SARSA} and \emph{Q-learning} algorithms,
thanks to the analogy with Markov decision processes established in Section~\ref{sec:transition}.

The reachability properties of stochastic digraphs can be also characterized by using the formulas~\eqref{eq:matrixEnt},
as done, e.g., in \cite[Prop.~1]{possieri2019mathematical}. However, the computational complexity of these
formulas is generally larger than the one using dynamic programming or stochastic optimization methods,
thus further motivating the interest in the results given in this technical note.

The following remark reviews different techniques to estimate the function $v_\pi(k,x)$
given the transition probabilities ${p}(j,r|i,a)$ and the reward function $r(i,a,j)$.

\begin{rem}\label{rem:solving}
In \cite{sutton2018reinforcement}, different methods are proposed to determine the function $v_\star(k,x)$
given the transition probabilities $p(j,r|i,a)$ and the reward function $r(i,a,j)$.
By the Bellman optimality equations, the function $v_\star(k,x)$ satisfies
\begin{equation}\label{eq:dynProg}
v_{\star}(k,x)=\max_a \sum_{j,r}p(j,r|i,a)\left(r+v_{\star}(k-1,j)\right).
\end{equation}
The dimension of the problem may hamper the analytical solution of the set of nonlinear equations~\eqref{eq:dynProg}.
Therefore, several algorithms, such as the \emph{policy iteration} and the \emph{value iteration}, have been proposed to
determine an approximate solution to~\eqref{eq:dynProg}. These algorithms have a worst case complexity
that is polynomial in the number of states and actions.
Other approaches, such as the {Q-learning} and the {SARSA} algorithms, determine an approximate solution
to~\eqref{eq:dynProg} from raw experience without a model of the MDP dynamics.
\end{rem}

\begin{rem}\label{rem:equiv}
The functions $\fR_{\Q}(k,x)$ and $\fV_{\Q}(k,x)$ can be also computed explicitly by leveraging the construction
used in the proof of \Cref{thm:bounds}. Namely, letting $m(k,x)$ and $\ell(k,x)$ be defined as in such a proof with $\mI_j$
replaced by $\Q$, by \cite{possieri2019mathematical}, for all $k\in\Zp$ and all $x\in\Zn$, one has that
\begin{align*}
\fR_{\Q}(k,x) & = m(k,x),&
\fV_{\Q}(k,x) & = \ell(k,x).
\end{align*} 
Thus, in principle, these bounds on the reachability probabilities can be computed
using the methods given in \cite{possieri2016weak,possieri2017asymptotic,possieri2018stochastic}, whose 
complexity is comparable to that of solving directly the dynamic programming equations~\eqref{eq:dynProg}.
On the other hand, by leveraging the correspondence between optimal state-value functions and reachability probabilities 
outlined in Propositions~\ref{prop:Wreach} and~\ref{prop:Wrec},
the other methods reviewed in \Cref{rem:solving} can be used, thus leading to smaller computational complexities,
especially in the case that one is interested to characterizing infinite-horizon reachability probabilities.
\end{rem}

\section{Case study: mobile agent-based epidemic models via stochastic digraphs\label{sec:disease2}}

The main aim of this section is to show how stochastic digraphs can be used to model diffusion phenomena in
populations of  mobile agents. With particular reference to epidemic processes, the results of Sections~\ref{sec:transition} and~\ref{sec:MDP} 
can be leveraged to determine upper and lower bounds for the temporal trends of the number of infected agents.
To exemplify the potentiality of our approach, we consider a susceptible-infected-removed (SIR) epidemic process \cite{brauer2012mathematical} that evolves over a mobile population of agents. Notably, we consider a group of $N$ agents that move over a graph $\M:=(\Z_{\varkappa},\W)$,
$\varkappa\in\Zp$. The state of the $i$-th agent is $\varphi_i=[\begin{array}{cc}
\sigma_i & \varpi_i
\end{array}]^\top$, where $\sigma_i$ is  $1$ for susceptible, $2$ for infected, or $3$ for recovered,
and $\varpi_i\in\Z_{\varkappa}$ denotes his position in the graph $\M$.
Letting $\N^o(\varpi)$ be the map of the out-neighborhoods of $\varpi\in\Z_{\varkappa}$ in
the graph $\M$, one may therefore describe the motion as
\begin{equation*}
\varpi_i^+\in \N^o(\varpi_i),\quad i=1,\dots,N,
\end{equation*}
where $\varpi_i^+$ denotes the position of the $i$-th agent in the graph $\M$ at the next time step.
Each susceptible agent can transition with probability $\alpha \in [0,1]$ to the infected state if it shares the same position with an infected agent $j$, i.e.,
	$\varpi_i=\varpi_j$ for some $i\neq j$. Note that each agent can be infected by any other agent sharing the same position and that these random events are considered independent of each other. Finally, each infected agent can transition to the recovered state with probability $\beta\in[0,1]$.
Thus, in such a model, the contact graph among agents is time-varying and is influenced by the movement
of agents over the graph $\M$.
The next theorem shows that
it is possible to construct a stochastic digraph, named $\bm{\SIRM}$, representing exactly the
behavior of the agent-based SIR model.

\begin{thm}\label{thm:SIRM}
Let the motion digraph $\M$, the infection probability $\alpha$ and the recovery probability $\beta$ be given.
Then, there exists a stochastic digraph $\bm{\SIRM}$ such that its stochastic directed paths are in one-to-one
correspondence with the trajectories of the agent-based SIR model with mobile agents.
\end{thm}
\begin{proof}
Let
$\sigma=[\begin{array}{ccc}
\sigma_1 & \cdots & \sigma_N
\end{array}]^\top$ denote the stack vector of all node states
and let 
$\varpi=[\begin{array}{ccc}
\varpi_1 & \cdots & \varpi_N
\end{array}]^\top$ denote the stack vector of the positions of the agents.
The updates of the state $\sigma_i$ are modeled by the stochastic difference inclusion
$\sigma_i^+\in G_i(\sigma,\varpi,u_{i},v_{i})$,
where 
\begin{multline*}
G_i(\sigma,\varpi,u_i,v_i)=\\
\begin{cases}
1,&
\begin{array}{l}
\hspace{-1ex}\text{if }\sigma_i=1\\
\land\left(\bigwedge_{\substack{
j=1\\j\neq i
}}^N\lnot(\varpi_i=\varpi_j\land\sigma_j=2\land [u_{i}]_j=2)\right),
\end{array}\\
2,&\begin{array}{l}
\hspace{-1ex}\text{if }\sigma_i=1\\\
\land\left(\bigvee_{\substack{
j=1\\j\neq i
}}^N(\varpi_i=\varpi_j\land \sigma_j=2\land [u_{i}]_j=2)\right),
\end{array}\\
v_i+1, &\text{if }\sigma_i = 2,\\
3,&\text{if }\sigma_i=3,
\end{cases}
\end{multline*}
and $(\bm{u_i})_k:\Omega\rightarrow\{1,2\}^N$, $(\bm{v_i})_k:\Omega\rightarrow\{1,2\}$ 
are two sequences of i.i.d. random variables with \begin{subequations}
\label{eq:probInp}%
\begin{align}
\bP([(\bm{u_{i}})_k]_j=2)&=\alpha,&j=1,\dots,N,\\
\bP((\bm{v_{i}})_k=2)&=\beta,
\end{align}
\end{subequations}
for $i=1,\dots,N$. 
Hence, define the set-valued map
\begin{equation*}
S_i(\sigma,\varpi,u_i,v_i):=\left[\begin{array}{c}
G_i(\sigma,\varpi,u_i,v_i)\\
\N^o(\varpi_i)
\end{array}\right],
\end{equation*} 
describing the next values of the epidemic state and of the position of the $i$-th agent given the values attained by the random inputs
$u_i$ and $v_i$.
Thus, let $\bar{n}=(3\,\varkappa)^N$ and let $\bar{\Pi}:(\{1,2,3\}\times \varkappa)^N\rightarrow \Z_{\bar{n}}$ be a bijective function
mapping each vector in $(\{1,2,3\}\times \varkappa)^N$ into an integer in  $\Z_{\bar{n}}$. Define
 $\omega=[\begin{array}{ccccccc}
u_1^\top & v_1 & \cdots & u_N^\top & v_N
\end{array}]^\top
\in \{1,2\}^{N^2+N}$
and consider the map $\bar{Q}:\Z_{\bar{n}}\times \{1,2\}^{N^2+N}\rightarrow\Z_{\bar{n}}$,
\begin{equation}\label{eq:mapG2}
\bar{Q}(x,\omega)=\bar{\Pi}\left(\left[\begin{array}{c}
S_1(\bar{\Pi}^{-1}(x),u_{1},v_{1})\\
\vdots \\
S_N(\bar{\Pi}^{-1}(x),u_{N},v_{N})
\end{array}\right]\right),
\end{equation}
where $\bar{\Pi}^{-1}$ is the inverse map of $\bar{\Pi}$.
Therefore, since the set $\{1,2\}^{N^2+N}$ is diffeomorphic to $\Z_{\bar{\err}}$, with $\bar{\err}=2^{N^2+N}$,
letting $\V=\Z_{\bar{n}}$,
the stochastic directed paths of the stochastic digraph $\bm{\SIRM}=(\V,\bm{\E},\mu(\cdot))$ corresponding to the map
$\bar{Q}(x,\omega)$ given in~\eqref{eq:mapG2} with the distribution $\mu(\cdot)$ induced by the sequences
of i.i.d. random variables $\bm{u_1},\bm{v_1},\dots,\bm{u_N},\bm{v_N}$ 
satisfying~\eqref{eq:probInp} are in one-to-one correspondence with the trajectories of the agent-based SIR model
with mobile agents.
\end{proof}

In the following example, we show how the stochastic digraph $\bm{\SIRM}$ together with the
 reinforcement learning techniques of \Cref{sec:MDP} can be used to determine upper and lower bounds 
 on the cumulative number of infected agents.
 
\begin{ex}\label{ex:SIRM}
Consider the agent-based SIR model with $N=3$, $\alpha=0.7$, $\beta=0.3$, and $\M$ as in \Cref{fig:mot}.

\begin{figure}[htb!]
\centering
\resizebox{0.15\textwidth}{!}{
\input{motionDigraph.tex}
}
\caption{Graph $\M$ used in \Cref{ex:SIRM}.\label{fig:mot}}
\end{figure}
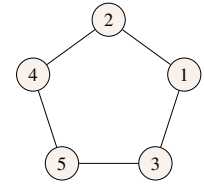

Using the construction in \Cref{thm:SIRM}, we determine the stochastic
 digraph $\bm{\SIRM}$ representing the agent-based SIR model.
 Then, using the technique given in \Cref{ssec:MDP}, we compute the transition probabilities $p(j|i,a)$ between the
 nodes of $\bm{\SIRM}$. Note that, in this setting, each node $i\in\V$ corresponds to a different state vector
 $\phi=[\begin{array}{ccc}
 \phi_1^\top & \cdots & \phi_N^\top
 \end{array}]^\top\in(\{1,2,3\}\times \{1,\dots,\varkappa\})^N$, and an action $a\in\A$ represents 
a motion law of all the agents. 

The upper and lower bounds on the cumulative number of infected agents can be determined using
the techniques described in \Cref{sec:MDP}.
Namely, let $\vartheta:\V\rightarrow\Zp$ be a map such that $\vartheta(i)$ is the number of agents that are either infected
or recovered in the state corresponding to node $i$. Hence, letting 
$\overline{r}(i,a,j)=\vartheta(j)-\vartheta(i)$ ($\underline{r}(i,a,j)=\vartheta(i)-\vartheta(j)$),
the optimal state-value function $\overline{v}_{\star}(i)$ ($\underline{v}_{\star}(i)$) is such that 
$\overline{v}_{\star}(i)+\vartheta(i)$ ($\vartheta(i)-\underline{v}_{\star}(i)$)
is an upper  (lower) bound on the cumulative number of infected agents.

Furthermore, the proposed framework can be used to compute the expected cumulative number of infected agents
in the case that the agents move on the graph uniformly at random, i.e., the next position of the $i$-th agent
is selected uniformly at random in $\N^o(\varpi_i)$,
Namely, letting $P$ be the matrix whose $(i,j)$-th entry is $p_{i,j}=\frac{1}{\vert \A \vert}\sum_{a\in\A}p(j|i,a)$,
which is the transition matrix of the Markov chain corresponding to a uniform random walk,
letting $\phi_0$ be the initial state of the agent-based SIR model, and letting $\Theta=[\begin{array}{ccc}
\vartheta(1) & \cdots & \vartheta(\bar{n})
\end{array}]^\top$, 
the expected cumulative number of infected agents at time $k$ is 
\begin{equation}\label{eq:varsigma}
\varsigma_k = \be_{\Pi(\phi_0)}\,P^k\,\Theta.
\end{equation}

Letting $\phi_0=[\begin{array}{ccccccc}
2 & 1 & 1 & 1 & 1 & 2
\end{array}]^\top$ the SARSA, and the Q-learning algorithms \cite{sutton2018reinforcement}
have been used to estimate $\underline{v}_{\star}(\Pi(\phi_0))$ and $\overline{v}_{\star}(\Pi(\phi_0))$.
Both algorithms have been used with learning rate equal to $0.1$
over $10^4$ episodes and employing an $\epsilon$-greedy policy with $\epsilon=0.1$.
The SARSA algorithm provided the estimates 
$\hat{\underline{v}}_{\star}^{\mathrm{SARSA}}(\Pi(\phi_0))=1.4978$ and 
$\hat{\overline{v}}^{\mathrm{SARSA}}_{\star}(\Pi(\phi_0))=-0.7011$,
whereas the Q-learning algorithm provided the estimates 
$\hat{\underline{v}}_{\star}^{\mathrm{Q}}(\Pi(\phi_0))=1.4922$ and 
$\hat{\overline{v}}^{\mathrm{Q}}_{\star}(\Pi(\phi_0))=-0.6986$.
\Cref{fig:expectedTotalInfected} illustrates the bounds obtained using these optimal state value functions,
the values $\varsigma_k$ gathered using~\eqref{eq:varsigma}, and
the results of a Monte Carlo simulation over $10^5$ samples
of the agent-based SIR model.

\begin{figure}[htb!]
\centering
\includegraphics[width=0.8\columnwidth]{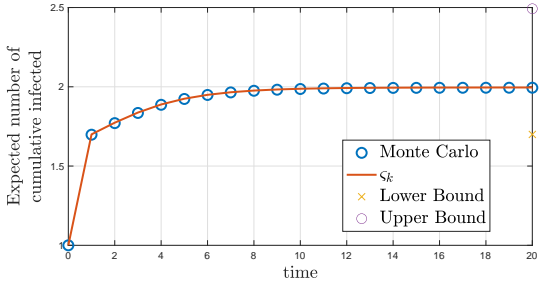}
\caption{Expected number of infected agents in \Cref{ex:SIRM}.\label{fig:expectedTotalInfected}}
\end{figure}

As shown in \Cref{fig:expectedTotalInfected}, the expected cumulative number of infected agents 
determined using~\eqref{eq:varsigma} and running  a Monte Carlo simulation are in agreement.
Furthermore, the values $\vartheta(i)-\underline{v}_{\star}(i)$ and $\vartheta(i)+\overline{v}_{\star}(i)$ are
asymptotic lower and upper bounds on the expected cumulative number of infected agents.
\end{ex}

Although in this section the attention has been focused
on SIR models, the proposed techniques can be easily extended to deal with other classes
of epidemic models \cite{possieri2019mathematical,brauer2012mathematical}.

\section{Conclusions}

Weak and strong reachability probabilities of a set for a stochastic digraph have been studied in this technical note.
The available state-of-the art techniques are characterized by a heavy computational burden that makes them unsuitable to efficiently tackle large graphs. In this technical note, we leverage techniques borrowed from the reinforcement learning domain to alleviate such a burden. Toward this aim,  we first showed that the transition probabilities in stochastic digraphs
can be modeled via a difference inclusion, which can be equivalently represented as
a Markov decision process. Hence, we have discussed how to select the reward function
 to determine upper and lower bounds for such transition probabilities
in the stochastic digraph. Crucially, we have formulated the transition probabilities as a function of an action space that depends on the current state, enabling a significant reduction of the problem size and paving the way to the definition of an efficient algorithm for their computation.

Notably, we discussed how stochastic digraphs can be used to model epidemic spreading 
and how our results apply to this framework. 
Our results are entirely general, and can be applied without any limitation to stochastic graphs of different nature, with applications in 
sensor networks, biological circuits, chemical reactions, and power grids.

\bibliographystyle{ieeetr}
\bibliography{biblio}

\end{document}

%% file: motionDigraph.tex
\begin{tikzpicture}[scale=0.5]

\node[circle,draw,black,fill = brown!10!white] (1) at (2.85317,0.927051) {1};
\node[circle,draw,black,fill = brown!10!white] (2) at (0,3) {2};
\node[circle,draw,black,fill = brown!10!white] (3) at (1.76336,-2.42705) {3};
\node[circle,draw,black,fill = brown!10!white] (4) at (-2.85317,0.927051) {4};
\node[circle,draw,black,fill = brown!10!white] (5) at (-1.76336,-2.42705) {5};

\draw (1)--(2);
\draw (2)--(4);
\draw (4)--(5);
\draw (5)--(3);
\draw (1)--(3);

\end{tikzpicture}